\documentclass[fleqn,10pt]{wlscirep}
\usepackage{multicol,multirow}
\usepackage[utf8]{inputenc}
\usepackage[T1]{fontenc}
\usepackage[most]{tcolorbox}
\usepackage{array}
\usepackage{tabularx}
\usepackage{placeins}
\usepackage{fvextra}
\usepackage{booktabs}

\newtcolorbox{cliniquote}{%
  colback=gray!5,
  colframe=gray!55,
  boxrule=0pt,
  leftrule=2pt,
  arc=0pt,
  outer arc=0pt,
  left=10pt, right=10pt,
  top=5pt, bottom=5pt,
  before skip=4pt, after skip=4pt,
}
\newtcolorbox{agentprompt}[1]{%
  colback=gray!5,
  colframe=gray!55,
  colbacktitle=gray!22,
  coltitle=black,
  fonttitle=\large\bfseries,
  title={#1},
  boxrule=0pt,
  leftrule=2pt,
  titlerule=0pt,
  arc=0pt,
  outer arc=0pt,
  left=10pt,
  right=10pt,
  top=6pt,
  bottom=6pt,
  toptitle=4pt,
  bottomtitle=4pt,
  lefttitle=10pt,
  righttitle=10pt,
  before skip=6pt,
  after skip=6pt,
  width=\linewidth,
  breakable,
}
\title{From Clinical Intent to Clinical Model: Autonomous Coding-Agents for Clinician-driven AI Development}
\author[1]{Zihao Zhao}
\author[1]{Eneko Cornejo Merodio}
\author[2]{Wenya Zhao}
\author[1]{Frederik Hauke}
\author[3]{Wensa Peng}
\author[4]{Xixi Song}
\author[1]{Maike Bode}
\author[1]{Juliana de Castilhos}
\author[5]{Mathis Bode}
\author[6,7,8]{Jakob Nikolas Kather}
\author[1]{Sven Nebelung}
\author[1,*]{Daniel Truhn}
\affil[1]{Department of Diagnostic and Interventional Radiology, University Hospital Aachen, Aachen, Germany}
\affil[2]{Department of Radiology, Nagoya City University Graduate School of Medical Sciences, Nagoya, Japan}
\affil[3]{CyberKnife Center, Huashan Hospital, Fudan University, Shanghai, China}
\affil[4]{Shanghai Geriatric Medical Center, Zhongshan Hospital, Fudan University, Shanghai, China}
\affil[5]{Juelich AI Factory, Forschungszentrum Jülich, Jülich, Germany}
\affil[6]{Else Kr\"oner Fresenius Center for Digital Health, TUD Dresden University of Technology, Dresden, Germany}
\affil[7]{Department of Medicine I, Faculty of Medicine and University Hospital Carl Gustav Carus, TUD Dresden University of Technology, Dresden, Germany}
\affil[8]{Department of Medical Oncology, National Center for Tumor Diseases (NCT), Heidelberg University Hospital, Heidelberg, Germany}

\affil[*]{dtruhn@ukaachen.de}

\begin{abstract}
\noindent\textbf{BACKGROUND}\quad Developing clinically useful AI requires collaboration between clinicians and AI developers, yet this process often involves time-consuming communication and potential misalignment. 
Coding agents may help bridge this gap. They can understand human requests, turn requests into working pipelines, and improve them iteratively. 

\noindent\textbf{METHODS}\quad We present a prototype that lets clinicians drive AI development directly via natural-language interactions.
The system turns a clinician request into an executable workflow through semantic parsing, task initialization, and iterative code tuning, using Claude Opus 4.6 as the underlying coding agent.
We evaluated this prototype across 12 tasks, including straightforward requests focused solely on model performance metrics and more complex requests.
The test set was held out from the beginning and was never exposed to the coding agent.
An exploratory human evaluation was conducted with six clinicians who completed a questionnaire rated on a five-point Likert scale after interacting with the system.

\noindent\textbf{RESULTS}\quad Across 12 tasks, the prototype autonomously developed AI models with competitive performance. 
Notably, clinicians' practical knowledge helped reduce shortcut reliance (partial correlation from 0.53 to 0.07) in pneumothorax classification. 
Drain-free AUROC improved in every cohort and the clinician-driven model achieved stronger lesion localization than medical foundation models across all five external cohorts. 
All six participants commented communication as efficient and agreed that the agent understood their clinical goals and can provide useful suggestions, but also saying extensive external testing is needed before clinical use of autonomously developed AI models. 

\noindent\textbf{CONCLUSIONS}\quad These findings suggest that coding agents could let clinicians take a more independent and active role in clinical AI development, reducing their reliance on AI developers, although further improvements are needed. 

\end{abstract}

\begin{document}

\flushbottom
\maketitle

\section*{Introduction}
\begin{figure}[!htbp]
    \centering
    \includegraphics[width=0.95\textwidth]{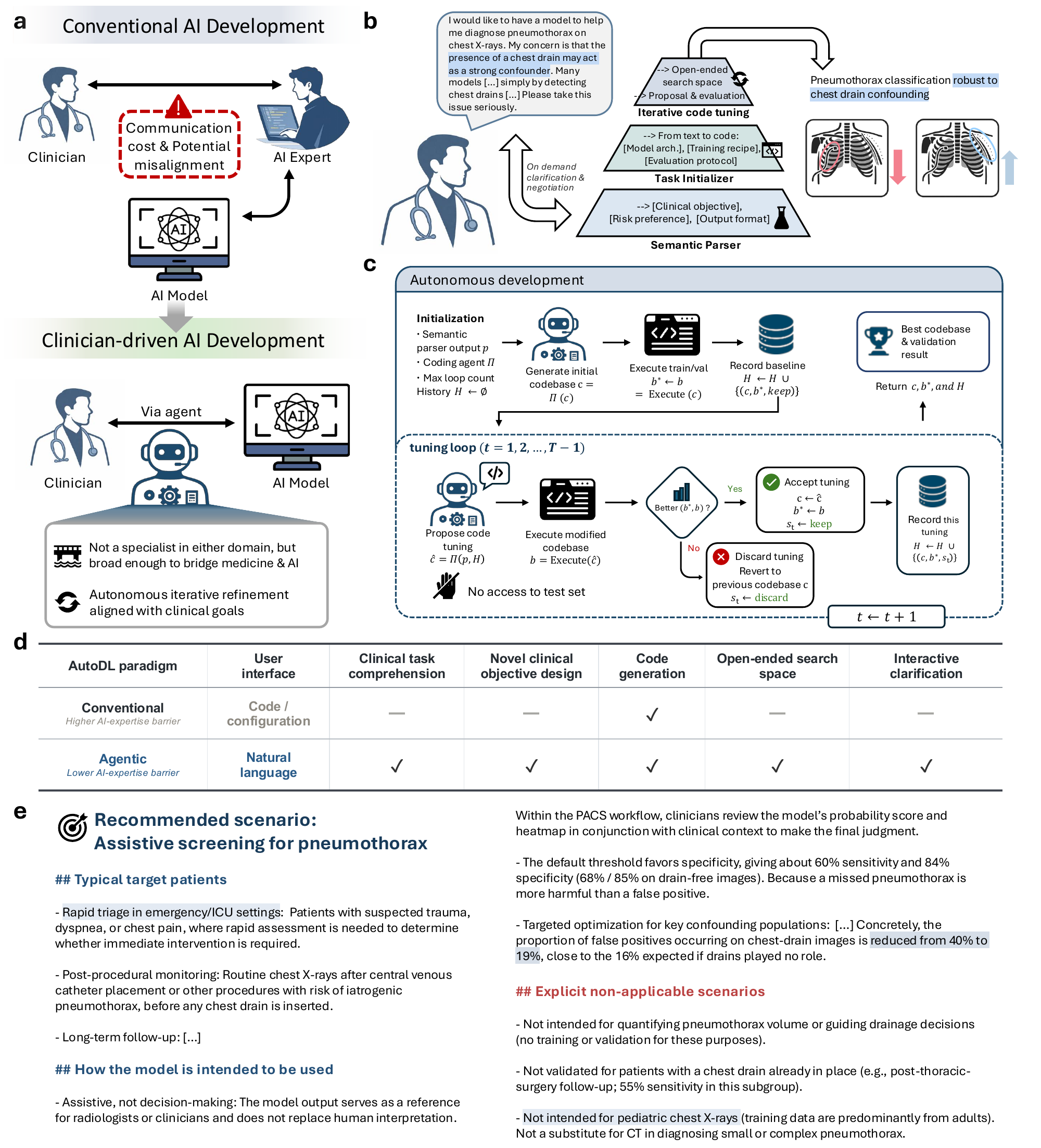}
    \caption{\textbf{Overview of clinician-driven AI development enabled by coding agents.} 
    \textbf{a,} Comparison between the conventional multi-party workflow and the proposed clinician-driven paradigm. Clinicians can guide model development directly through coding agents, reducing the need for communication overhead and aligning model optimization with clinical preference.
    \textbf{b,} Pipeline of clinician-driven AI development: a natural-language request is translated into executable code and iteratively refined to satisfy clinical objectives, illustrated here by pneumothorax classification robust to chest-drain confounding. 
    \textbf{c,} Iterative code tuning loop of the autonomous developer, where modifications are accepted only if they improve a predefined validation objective, without access to the test set. 
    \textbf{d,} Comparison between conventional AutoDL and clinician-driven agentic AutoDL.
    \textbf{e,} Example of user guidance generated by the agent, including recommended use scenarios, target populations, and known limitations.
    }
    \label{fig:overview}
\end{figure}

Development of artificial intelligence (AI) for medicine has traditionally relied on close collaboration between clinicians and specialized AI teams~\cite{ghassemi2020review,mongan2021rsna,galsgaard2022artificial}. 
In this workflow, clinicians specify the clinical problem and its intended use, while AI engineers make decisions about data preparation, model architecture, training, and evaluation. Despite the progress enabled by such collaborations, translating clinical requirements into technical specifications remains time-consuming and communication-intensive~\cite{wysocki2023assessing,gisselbaek2025bridging,bienefeld2023solving,sokol2025artificial}.
Many experienced clinicians are not familiar with AI development~\cite{hwang2025translating,bold2026global} and may not know what AI models can and cannot do.
Hence, multiple rounds of discussion are often required, yet this does not necessarily ensure faithful alignment between the two sides~\cite{bienefeld2023solving,sokol2025artificial}.

Recent advances in coding agents may change this long-standing paradigm~\cite{yang2024swe,grishina2025fully,openai2025codex,anthropic2025claudecode,tayebi2024large,zhao2026can,lu2026towards}. Large language model (LLM)-based agents are increasingly capable of writing code, configuring experiments, debugging failures, and iteratively refining solutions with limited human intervention~\cite{yang2024swe,grishina2025fully,openai2025codex,anthropic2025claudecode}. More recently, the emergence of ``AI scientist'' systems has further suggested that complex workflows could be partially automated through natural-language interaction alone~\cite{tayebi2024large,lu2026towards,gottweis2026accelerating,aygun2026ai}. For clinical AI, this suggests that clinicians may no longer need to rely entirely on specialized AI developers. Instead, they could describe clinical tasks in natural language and directly guide the model development process~\cite{kather2024large}. Although current LLM-based agents can not yet match top-tier domain specialists in either medicine or deep learning~\cite{takita2025systematic,jimenez2024swebench}, their broad cross-domain knowledge allows them to function as an effective bridge between clinicians and AI development~\cite{wang2025perspective}. As illustrated in Figure~\ref{fig:overview}a, such a shift could reduce the communication costs. By allowing clinicians to guide model development directly, it may also help preserve the original clinical objectives throughout the development process.

 In this study, we present a working prototype of what we call \emph{clinician-initiated, autonomously developed} clinical AI: the clinician specifies the task and its clinical priorities in natural language, and the system carries out model development from there. The system turns this request (hereafter a ``clinician request'') into an executable workflow in three stages (Figure~\ref{fig:overview}b). Semantic parsing first converts the clinical intent into a structured task representation. This representation is meant for clinicians to verify whether their requests have been clearly interpreted. Then, the task initialization module maps it to a model architecture, training recipe, and evaluation protocol. Iterative code tuning subsequently builds the model through code generation, debugging, and experimentation.The latter two constitute the autonomous development (Figure~\ref{fig:overview}c). Throughout this process, clinicians can clarify their requirements and negotiate trade-offs with the agent whenever needed.
 Compared with conventional automated deep learning (AutoDL)~\cite{elsken2019neural,he2021dints,zimmer2021auto}, which relies on predefined settings and search spaces, the coding agent-based approach offers greater flexibility in how clinical AI models are specified and developed (Figure~\ref{fig:overview}d). It can understand clinical requirements, customize task-specific objectives, and explore possible modeling solutions in an open-ended manner through natural-language interaction. Interactive feedback and clarification also make the development process more transparent than that of conventional AutoDL methods.
After model development, the agentic system also generates guidance for clinical use (Figure~\ref{fig:overview}e). It describes where the model may be useful, how it can be integrated into clinical workflows, and when it should not be used. We do not claim that this framework removes all barriers to clinical AI development. Rather, we present it as a proof of concept that recent progress in coding agents has made clinician-driven model development technically plausible.

We evaluated this approach across 12 tasks, including straightforward requests focused solely on model performance metrics and more complex requests involving specific requirements for model design or intended use. We further investigated the value of placing clinicians at the center of AI development, showing how their domain expertise can bring substantial improvements to model performance when translated into technical solutions by coding agents. In turn, we explored how coding agents can help clinicians refine their requirements and modeling decisions through interactive clarification and feedback. Finally, an exploratory user study revealed encouraging clinician-reported experiences, particularly in development independence, communication efficiency, and the usefulness of agent feedback. However, participants remained hesitant about using the resulting models in clinical practice. These findings suggest that coding agents could give clinicians greater independence in clinical AI development, although further improvements in coding agents are needed to develop models that clinicians can confidently rely on in clinical practice.

\section*{Methods}\label{sec:method}

\subsection*{Data curation}
Both public and private datasets were used in this study. Whenever a dataset was used as a development cohort and provided official training and test splits, those splits were adopted directly. If no official validation split was available, 20\% of the original training data were reserved for validation. Splitting was performed at the patient level whenever identifiers were available and otherwise at the sample level. The class balance was preserved during the training and validation splitting process. These constraints were made to prevent data leakage and to ensure that validation remained genuinely separate from training.
For datasets serving as external test cohort, the entire dataset was used for evaluation.
Please refer to Table~\ref{tab:splits-overview} for splitting details in each experiment.

\subsection*{Evaluation protocol}
For each task derived from clinician's request, its specific evaluation function was determined during semantic parse and task initialization, and then kept fixed throughout the code refinement loop. Each task was associated with one primary validation metric, used exclusively for model selection, and a set of additional secondary metrics logged for interpretability and later statistical analysis. 
The direction of the primary metric (i.e., whether improvement corresponds to an increase or a decrease in its value) was explicitly specified at initialization. This design choice enhances the robustness~\cite{yang2025number,mahendra2025evaluating,gonzalezpumariega2026reliabilitycomputeruseagents} of the iterative code tuning process by formalizing the improvement criterion, ensuring that performance comparisons across runs remain consistent and reliably reflect true improvements.

During the pipeline of clinician-driven AI development, the test set was held out from the beginning and was never exposed to the coding-agent. It only has access to training and validation splits in this process. The test set performance of the corresponding clinical AI model was evaluated only after the end of 30-round refinement. Reported test-set results therefore reflect post hoc evaluation, preventing coding-agent from optimizing code tuning indirectly based on test-set scores.

To quantify uncertainty and compare the initial and final refined models, we used paired bootstrap resampling~\cite{efron1992bootstrap}. Specifically, 10{,}000 bootstrap samples were drawn to estimate 95\% CIs~\cite{hazra2017using} and to test for differences between models. No correction for multiple comparisons was applied.

\subsection*{Agent configuration and context design}
The proposed autonomous prototype is implemented as a single coding agent (Claude Opus 4.6) operating within a structured execution harness~\cite{openai_harness_engineering}.
All experiments were conducted using Claude Code command-line interface. The agent was used in a continuous, stateful interaction to perform semantic parsing, task initialization, and autonomous development within a single execution context, rather than as separate components in a multi-agent system.

The agent was executed under the default configuration of the Claude Code interface. No manual adjustments were made to decoding parameters (such as temperature or sampling strategy), and no explicit control of randomness was imposed. 
Iteration budget, acceptance rule, evaluation function, and data-handling constraints was predefined and held fixed across all experiments.

All task constraints and experimental rules defined in below subsections were provided to the agent in the form of project-level context files, rather than being embedded directly as prompts, so that these constraints could be revisited throughout a long-horizon interaction while remaining explicit and auditable~\cite{zhou2026externalization,zhou2026mem}. In particular, three core files, \texttt{program.md}, \texttt{FINALIZE.md}, and \texttt{Parser.md}, defined complementary aspects of the system. The former two specified the overall experiment workflow, including iteration structure, data-splitting rules, model-selection criteria. How the coding agent should pack and send usage manual, inference examples, and all other requested files to the user was also defined in \texttt{FINALIZE.md}. At the same time, the latter defined the schema for translating clinician requests into structured task specifications.
This design ensured that the operational rules governing model development were explicitly defined and externally accessible, rather than implicitly encoded within prompt instructions.

When initiating a new experiment, a short instruction will be concatenated to the end of the clinician's request, directing the agent to inspect \texttt{program.md} and begin experiment setup. The agent then autonomously parsed the request, consulted the available context files, and initialized a new experiment without further manual intervention.
Detailed specifications defining the agent behavior at different stages are provided in Section~\ref{sec:appendix_prompt}.

\subsection*{Parsing and initialization}

Each experiment began with a clinician-style natural-language request, which was converted into a structured task specification by the agent. The specification was represented in a JSON format and included the task type, clinical objective, risk preference, input modality, expected inference output, evaluation metrics, and a primary metric with an explicitly defined optimization direction. 
To improve the reliability of this translation, the parsing stage was guided not only by the raw clinician request but also by embedded AI expert knowledge provided in textual form. These additional constraints were designed to reduce ambiguity in cases involving asymmetric risk preferences (for example, prioritizing sensitivity to avoid missed disease).
The resulting structured specification served as the interface between the clinician request and downstream model development. It was subsequently instantiated as a minimally executable codebase that satisfied predefined constraints.

\subsection*{Iterative code tuning}

Following initialization, the agent iteratively refined the codebase. The total number of refinement iterations (runs) was set to 30, with each run has a 30 minutes time budget. Two NVIDIA L40S GPUs were provided by default. Hence, a complete iterative tuning process corresponded to a nominal training budget of around 30 GPU-hours. The agent was allowed to stop early, but only after completing at least 15 iterations. At each run, a single code modification was proposed, executed through a complete training--validation cycle based on the predefined primary metric. The codebase was updated only if the modification produced a strict improvement in the primary validation metric. Modifications yielding equal or worse performance were discarded. Runs that failed to complete due to out-of-memory errors or excessive runtime were treated as failed attempts and hence discarded.  
Although only performance-improving modifications were incorporated into the codebase, all modifications were recorded and remained accessible to the agent throughout the process. This ensured that subsequent iterations could condition on both successful and unsuccessful prior attempts.
The final training configurations reached by the agent for the tasks are available in our
\href{https://github.com/zhaozh10/clinical-automata}{GitHub repository}.

\subsection*{Exploratory human evaluation}

We conducted the exploratory user study via online web demo. Participants interacted with the coding agent remotely through the web-based interface (Supplementary Figure~\ref{fig:web_demo}), without requiring programming knowledge, command-line access, or local computational resources. The interface provided preloaded medical-imaging datasets, accompanied by dataset descriptions and example requests. Participants could explore these datasets, consult the agent for recommendations, and formulate their own clinical AI development tasks. The agent then interpreted the request and returned a structured task specification, which clinicians could review, clarify, or revise through further dialogue. After initialization and getting clinician's approval, model development then proceeded autonomously on remote computational resources. Participants were notified by email upon completion. During the user study, each participant could access only the datasets they were authorized to use through the web-based interface. Publicly available datasets were offered to all participants, whereas the in-house paediatric hip radiograph cohort was visible only to participants from the radiology department of University Hospital RWTH Aachen who were authorized to access it.

We recorded the waiting time for AI model initialization during their session, which was defined as the period from the start of the session to the first working pipeline, excluding intervals when the agent was waiting for clinician's response. Hence, later iterative code tuning process was not included, and this definition should not be interpreted as total development time cost.

Following their experience with the system, the six clinicians completed a questionnaire comprising ten statements rated on a five-point Likert scale, from strongly disagree to strongly agree. The questionnaire assessed the agent's understanding of clinical intent, usefulness of its feedback, communication efficiency, perceived transparency, development independence, perceived control, acknowledgment of limitations, trust in clinical use, and overall satisfaction. One reverse-worded item assessed difficulty communicating clinical intentions. Participants could also provide open-ended feedback. 
All participants provided informed consent, and their questionnaire responses were analyzed anonymously. Given the exploratory nature of the study and its small sample size, questionnaire responses were analyzed descriptively as counts and proportions, without inferential statistical testing.

\begin{figure}[!tb]
    \centering
    \includegraphics[width=\textwidth]{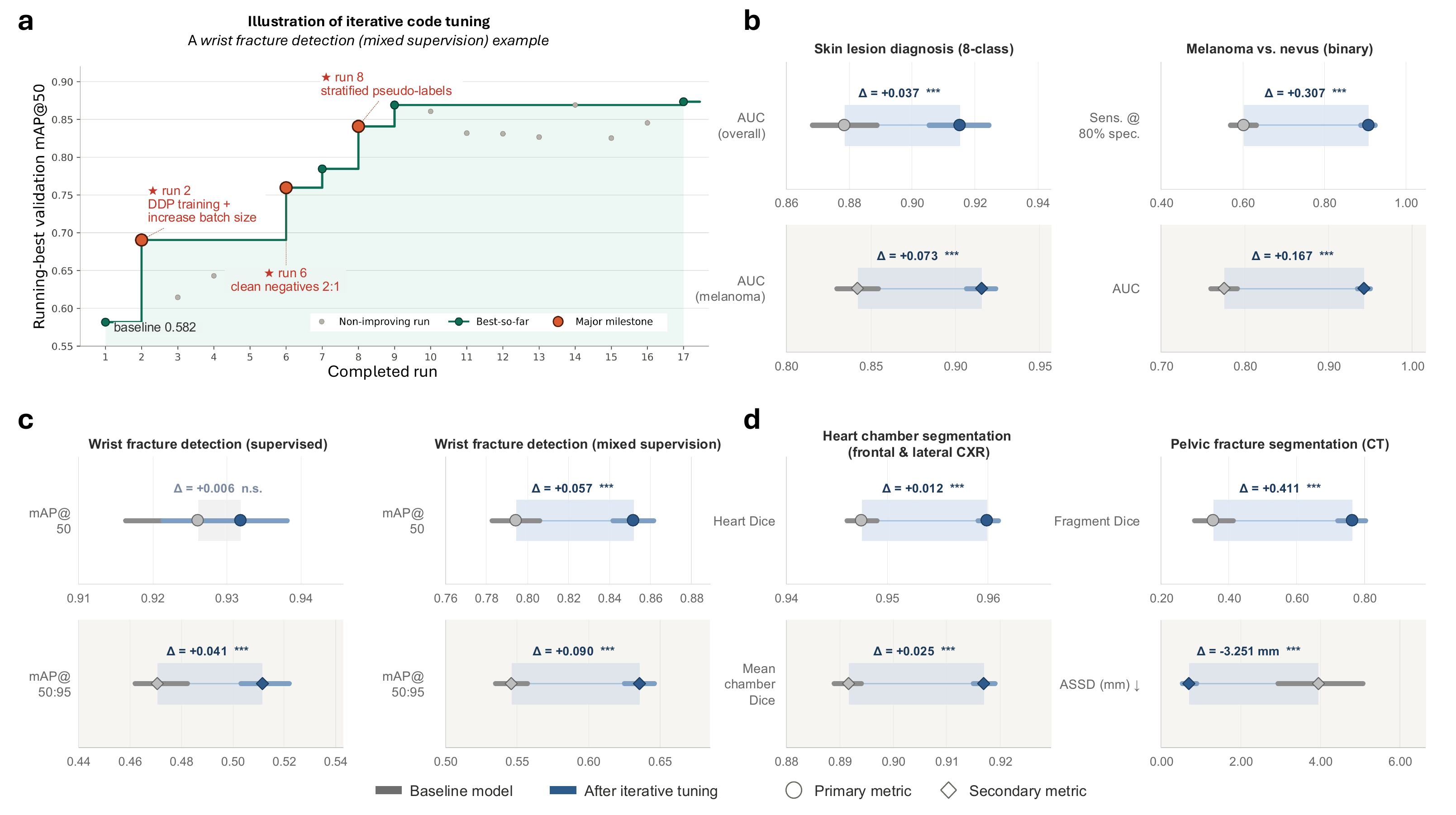}
    \caption{\textbf{Coding agents implement autonomous model development for straightforward tasks.}
    \textbf{a,} Representative development process for wrist fracture detection with mixed supervision. After baseline initialization, the coding agent autonomously proposes and validates code changes, and only preserves changes bringing improvment. The plot shows the performance of each individual experimental runs and illustrating the best validation performance so far.
    \textbf{b-d,} Comparison of model performance(95\% CIs from 10,000 paired bootstrap samples) before and after iterative code tuning across multiple straightforward medical imaging tasks. Statistical significance was assessed using paired bootstrap tests.
}
\label{fig:simple_request}
\end{figure}
\section*{Results}
We first tested whether coding agents could build a baseline model and improve it progressively toward the specified objectives on six straightforward tasks. Then, we examined five of the six complex tasks to show how coding agents can handle real-world requests that go beyond standard performance metrics. The remaining complex task was specifically selected to illustrate the benefits clinicians' expertise could bring to coding agents.
We also explored how could coding agents, in turn, support clinicians. Finally, we assessed clinicians' experiences with the agentic system.

\subsection*{Autonomous and progressive development of clinical AI models}
Figure~\ref{fig:simple_request} show the evaluatation across six straightforward tasks, covering classification~\cite{cassidy2022analysis}, detection~\cite{nagy2022pediatric}, and segmentation~\cite{seibold_2026_20688438,liu_2026_19732767}. In these tasks, users only specified the medical image domain and a straightforward objective, for example:

\begin{cliniquote}
\textit{``I want an AI model to help diagnose skin lesions. Priority should be given to melanoma.''}
\end{cliniquote}
\begin{cliniquote}
\textit{``I want to have an AI model to help me distinguish melanoma from nevus. Please try your best to not miss a single case.''}
\end{cliniquote}
\begin{cliniquote}
\textit{``I need a tool that can look at wrist X-rays and help me find fractures. Ideally, it should mark suspicious fractures on images during my workflow so they are less likely to be missed.''}
\end{cliniquote}

A wrist fracture detection task under mixed supervision setting shows how this process worked (Fig.~\ref{fig:simple_request}a).
We select this example because it resembles the real-world situation in which clinicians have a large number of images but only a small fraction of them are annotated with bounding boxes. Specifically, we limit the availbility of bounding boxes annotations for the given GRAZPEDWRI-DX dataset~\cite{nagy2022pediatric}. Only 5\% of the training images (425 samples) had access to fracture bounding-box annotations, while the remaining 95\% (7,637 samples) only had image-level labels indicating the presence of fractures. The clinician request was identical to that used in the fully annotated setting, without instructions on how to handle the issue of limited annotations.

Over the entire autonomous development process, the running-best validation mAP@50 rose from 0.582 to 0.874.
The agent first improved the basic setup through distributed training and a larger batch size.  It then reduced the share of fracture-free images in the training set from about ten per annotated positive to two, reasoning that most training steps were otherwise spent on negatives (run 6). Next, it used the detector trained in its previous run to generate pseudo-boxes on the weakly labeled fracture images, adding predictions with high confidence to the training set (runs 7--8). We notice that not all attempts helped. Attempts include larger input resolution, larger backbones, and several augmentation did not improve validation performance and were hence discarded. 

We compared the baseline and after-iterative-tuning models on held-out test sets across all six tasks (Figure~\ref{fig:simple_request}b-d). In eight-class skin lesion classification, overall AUC increased by 0.037, with a larger gain of 0.073 for melanoma AUC. In the binary melanoma-versus-nevus task, sensitivity at 80\% specificity increased from 0.602 to 0.909, while AUC increased from 0.775 to 0.942. Wrist fracture detection with mixed supervision gained 0.057 in mAP@50 and 0.090 in mAP@50:95. The fully supervised fracture task showed a smaller change in mAP@50 (+0.006), which was not statistically significant, although its mAP@50:95 improved by 0.041. Improvements in the primary metrics of the other five tasks were significant in paired bootstrap tests ($p<0.001$). The agent also improved heart segmentation on chest radiographs (heart Dice, +0.012) and pelvic fracture fragment segmentation on CT (fragment Dice, +0.411). Overall, these experiments show that the coding agent could move from a straightforward request to a working model and improve it progressively toward clinical objectives.

\begin{figure}[!htbp]
    \centering
    \includegraphics[width=\textwidth]{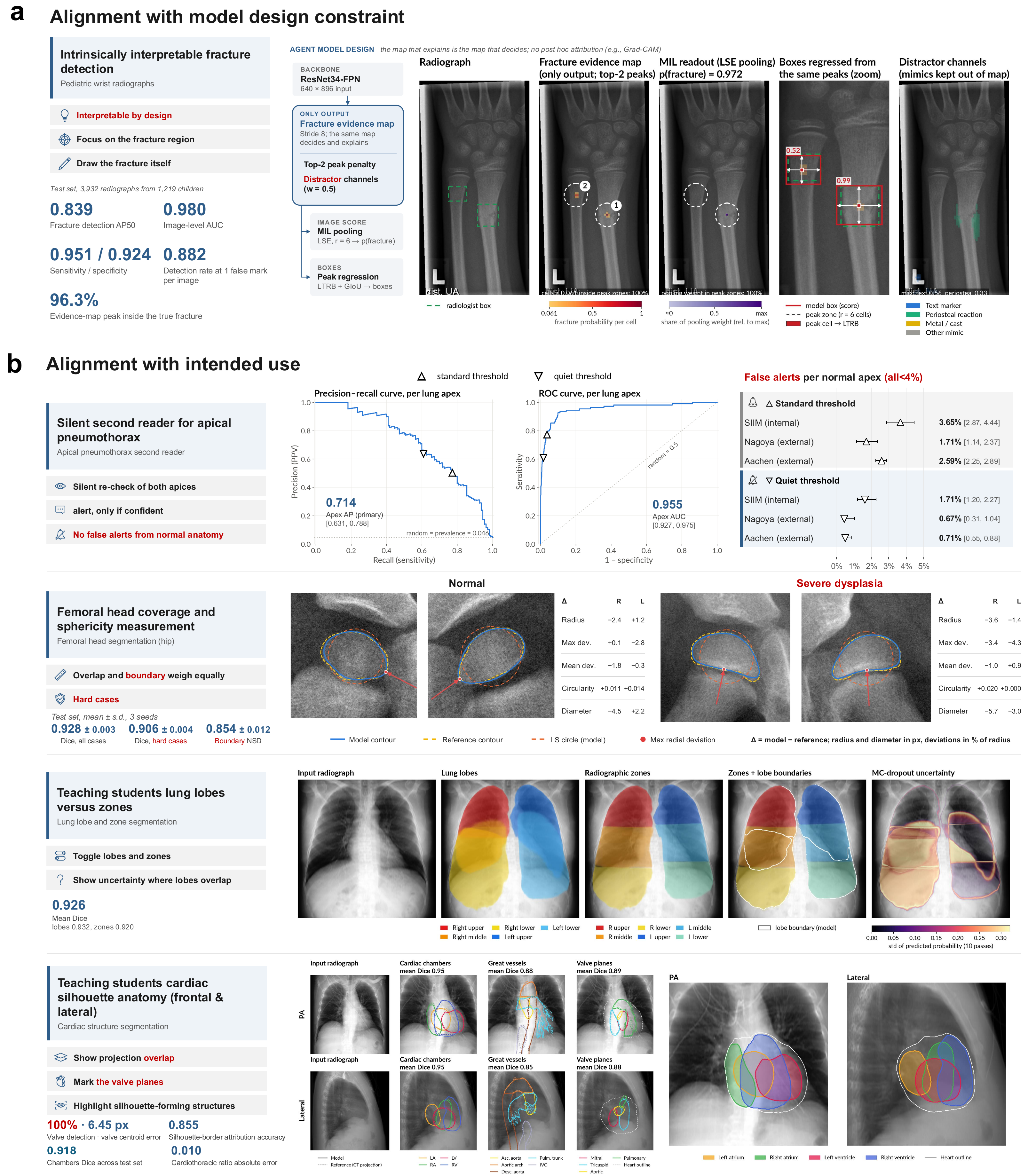}
    \caption{\textbf{Coding agents develop models aligned with complex, real-world clinical requirements.}
\textbf{a,} Alignment with model design constraints. In one real-world example, the clinician requests an intrinsically interpretable model in which the regions used for prediction also provide evidence for fracture localization.
The coding agent addresses this constraint through multi-instance learning and distractor prediction.
\textbf{b,} Alignment with intended use. Four real-world examples demonstrate how coding agents adapt model development to address different clinical needs. Quantitative results and representative predictions are shown.
}
\label{fig:complex_request}
\end{figure}
\begin{figure}[!htbp]
    \centering
    \includegraphics[width=0.90\textwidth]{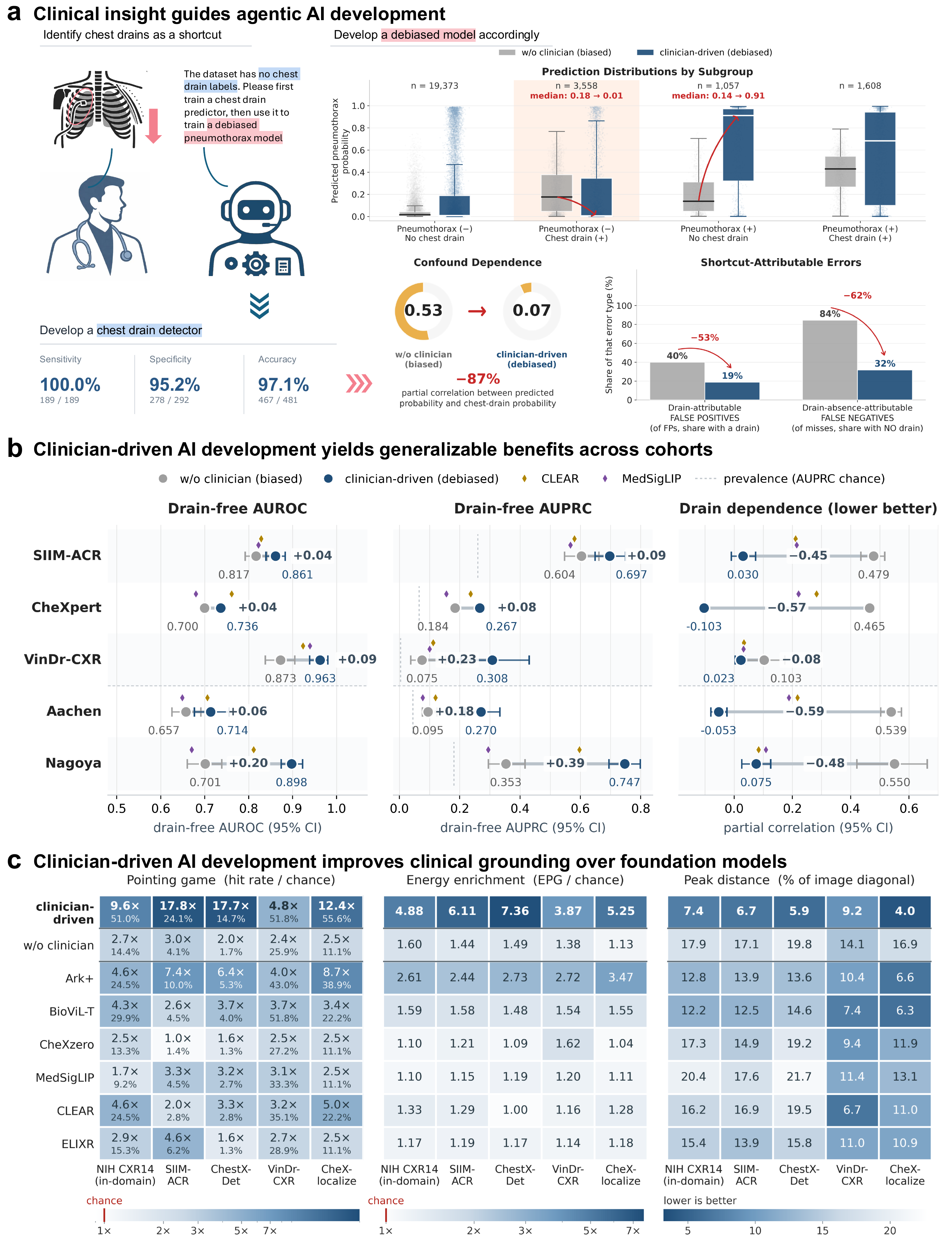}
    \caption{
    \textbf{Clinicians guide coding agents through clinical insights.}
\textbf{a,} The clinician identifies chest drains as a potential shortcut, and the coding agent proposes training a chest-drain detector
 to address missing annotations before developing a debiased pneumothorax model.
Internal testing shows remarkable reduction in confound dependence and shortcut-related errors.
\textbf{b,} Improvements in multiple shortcut-related metrics across cohorts demonstrate generalizable benefits, 
with CLEAR and MedSigLIP included as representative foundation model baselines.
\textbf{c,} Clinician-driven debiased model achieves better clinical grounding performance than medical foundation models
 across five cohorts.
}
\label{fig:why_clinician}
\end{figure}
\subsection*{Addressing complex clinician-specified requests beyond metrics}
We next put emphasis on five complex requests that go beyond simple performance metric requirements. These requests were made by \textbf{real clinicians} and concerned how a model should work or how it would be used.

In Figure~\ref{fig:complex_request}a, one clinician requested a wrist fracture detector that was interpretable by design. The clinician wanted the image regions driving the fracture prediction to also identify the fracture location, rather than relying on a separate post hoc explanation. To meet this requirement, the agent developed a model with a shared fracture evidence map, combining multi-instance learning~\cite{quellec2017multiple,yan2016multi} with distractor prediction~\cite{wang2023grenet,hazra2025evaluating}. The resulting model achieved a fracture detection AP50 of 0.839 and an image-level AUC of 0.980. In 96.3\% of evaluated fractures, the highest point on the evidence map fell inside the annotated fracture region. The model thus used the same evidence map to make predictions and localize fractures, as the clinician had requested.

Other requests focused on how the models would be used (Figure~\ref{fig:complex_request}b). For apical pneumothorax screening, a clinician requested a second reader that could identify suspicious findings while limiting unnecessary alerts. The agent hence developed a model with two operating thresholds, standard and quiet, with the latter one as the default recommendation and the first one as both a baseline and the other choice if clinicians needed. With the quiet threshold, false alerts per normal apex decreased from 3.65\% to 1.71\% on the internal SIIM cohort, from 1.71\% to 0.67\% on the external Nagoya cohort, and from 2.59\% to 0.71\% on the external Aachen cohort. ~\textbf{Notably, the quiet setting achieved false-alert rates below 1\% on both external cohorts, an encouraging result for a model intended to operate without disrupting routine clinical reading.} The model also achieved an average precision of 0.714 and an AUC of 0.955. 
In the second row of Figure~\ref{fig:complex_request}b, a clinician requested femoral-head segmentation models. 
The clinician emphasized that accurate boundaries are crucial, as even a slight contour error can significantly affect subsequent measurement results.
Based on this clinical goal, the agent defined a task-specific evaluation score that weight region overlap (Dice) and boundary agreement (normalized surface Dice, NSD) equally. The resulting model achieved a mean Dice of $0.928\pm0.003$ across all cases, $0.906\pm0.004$ on difficult cases, which were manually categorized by the clinician.
The last two  examples focused on medical education, further illustrating the flexibility of clinician-driven development.

\FloatBarrier
\subsection*{The value of clinician leadership in AI development}
Here, we investigated how clinicians' domain expertise could guide AI development and improve the resulting models. We studied pneumothorax classification on chest radiographs, where a clinician identified chest drains as a potential source of shortcut learning (left part of Figure~\ref{fig:why_clinician}a). Chest drains are often inserted after the diagnosis of pneumothorax, making their presence strongly associated. The clinician recognized this association as a potential shortcut reliance that might mislead AI models, hence making reducing chest-drain dependence an explicit request for the coding agent.
\begin{figure}[htbp]
    \centering
    \includegraphics[width=0.99\textwidth]{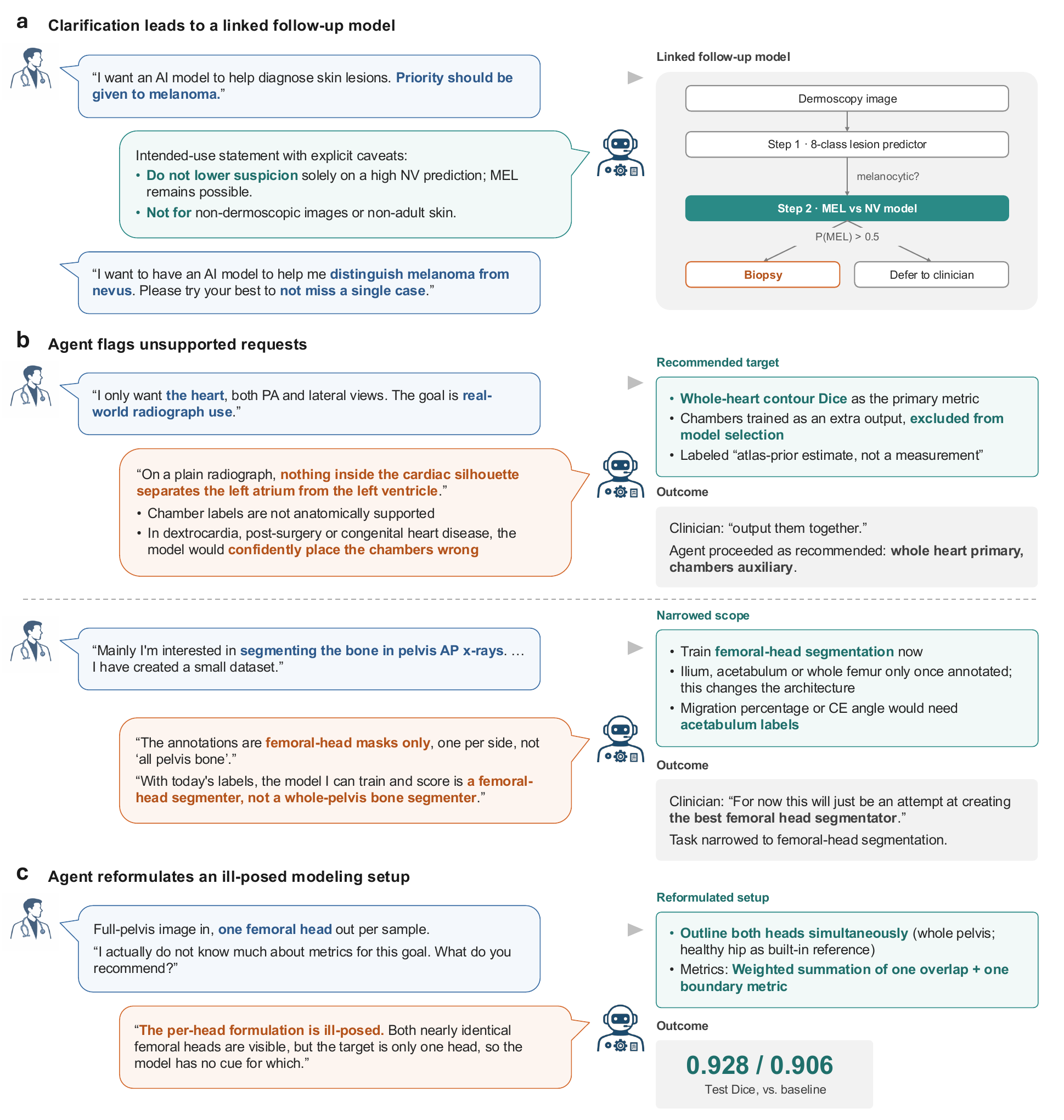}
    \caption{\textbf{Coding agents support clinicians through interactive feedback.}
    \textbf{a,} The agent clarifies clinical needs and suggests a linked follow-up model for dermoscopy diagnosis, emphasizing the importance of melanoma vs. nevus classification.
    \textbf{b,} The agent identifies limitations in clinical requests, recommending alternatives or more appropriate task definitions.
    \textbf{c,} The agent corrects poorly defined AI modeling setup in clinicians' requests through dialogue.
    }
\label{fig:feedback_interaction}
\end{figure}
The NIH ChestX-ray14~\cite{wang2017chestx} dataset lacked chest-drain annotations, so addressing this concern is necessary. The coding agent hence proposed training an auxiliary chest-drain detector using a small-scale public dataset~\cite{cheplygina2025augmenting}. The clinician approved this proposal, and the agent developed the detector and incorporated its predictions into the pneumothorax training pipeline. On the held-out chest-drain detection set, this auxiliary detector model achieved 100.0\% sensitivity and 97.1\% accuracy (Section~\ref{sec:appendix_drain}). The agent then used the predicted drain labels to guide pneumothorax model development. Its debiasing strategies included balancing training samples across pneumothorax and drain subgroups, adversarial training with gradient reversal~\cite{ganin2015unsupervised} to reduce the information about chest drains learned by the model, and regularization with mixup~\cite{zhang2018mixup} and label smoothing~\cite{szegedy2016rethinking}.

On the internal test cohort, the partial correlation between pneumothorax and chest-drain decreased from 0.53 to 0.07. At the same time, the percentage of chest-drain attributable false positives fell from 40\% to 19\%, while the drain-absent attributable false negatives fell from 84\% to 32\%. The prediction distributions (Top right in Figure~\ref{fig:why_clinician}a) also showed promising distribution shift. Pneumothorax-negative images with drains received lower disease probabilities, and pneumothorax-positive images without drains received higher probabilities. These findings indicate a marked reduction in dependence on the shortcut identified by the clinician.

The benefits extended across five test cohorts, including external datasets (Figure~\ref{fig:why_clinician}b). Compared with the model developed without clinician-guided debiasing, drain-free AUROC improved in every cohort. It increased from 0.817 to 0.861 on SIIM-ACR~\cite{siim-acr-pneumothorax-segmentation}, from 0.873 to 0.963 on VinDr-CXR~\cite{nguyen2022vindr}, and from 0.701 to 0.898 on the Nagoya cohort. Drain-free AUPRC also improved across cohorts, accompanied by lower measured chest-drain dependence. 
We further compared the models with two representative medical foundation models, CLEAR~\cite{han2026clear} and MedSigLIP~\cite{sellergren2026open}. CLEAR is a chest X-ray foundation model marked by resistance to shortcut features, while MedSigLIP is a general medical imageing foundation model trained on enormous data samples. 
Nevertheless, the clinician-driven model generally achieved higher drain-free AUPRC, together with lower drain dependence and comparable drain-free AUROC, than CLEAR. Its advantage was pronounced when compared with MedSigLIP.

In Figure~\ref{fig:why_clinician}c, we investigated the model's pneumothorax grounding ability against a broader set of medical foundation models, including Ark+~\cite{ma2025fully}, BioViL-T~\cite{bannur2023learning}, CheXzero~\cite{tiu2022expert}, MedSigLIP, CLEAR, and ELIXR~\cite{xu2023elixr}. The clinician-driven model achieved stronger lesion localization across all five evaluation cohorts, with particularly pronounced gains in pointing game~\cite{zhang2018top} and energy enrichment. Pointing game reached 17.8 times chance on SIIM-ACR and 17.7 times chance on ChestX-Det~\cite{lian2021structure}. The model also showed more concentrated attribution within annotated lesions and generally shorter distances between attribution peaks and lesion locations. These results indicate that the clinician-guided development process improved the model's ability to highlight radiographic regions relevant to pneumothorax.

These results highlight the value of clinician leadership in AI development. The clinician's understanding of a clinically important failure mode shaped the development objective and led to substantial improvements in model robustness, generalization, and lesion localization.

\FloatBarrier
\subsection*{Agent feedback supports clinician-led development}
For skin lesion diagnosis in Figure~\ref{fig:feedback_interaction}a, the agent recognized the clinical importance of distinguishing melanoma from nevus within a broader eight-class classification task. It then suggested developing an additional melanoma-versus-nevus classifier, which became part of the subsequent development plan. This interaction changed the proposed development plan rather than only changing a hyperparameter.
The agent also helped clinicians align their requests with the information available in the provided dataset (Figure~\ref{fig:feedback_interaction}b). When a clinician requested cardiac chamber segmentation from chest radiographs, the agent explained the limited visibility of individual chamber boundaries. It hence recommended using the whole-heart contour as the primary evaluation target. In another case, a clinician asked for broad pelvic bone segmentation, while the available annotations include only the femoral heads. The agent identified this mismatch and helped the clinician narrow the task to femoral-head segmentation.

A further example involved a clinician with limited AI experience. Specifically, the user requested segmentation of a single femoral head from a full-pelvis radiograph without specifying the target side (Figure~\ref{fig:feedback_interaction}c), indicating two independent segmentation models for two sides of the femoral head. The agent identified the ambiguity and suggested segmenting both femoral heads jointly in one single model instead . It also helped formulate an evaluation objective that considered both region overlap and boundary quality. 

Across these examples, we show that the agent did not blindly accept requests from clinicians. Instead, it could explain constraints and propose alternatives for the clinician to consider. The final clinical goals and decisions remained with the clinician.

\begin{figure}[htbp]
    \centering
    \includegraphics[width=0.99\textwidth]{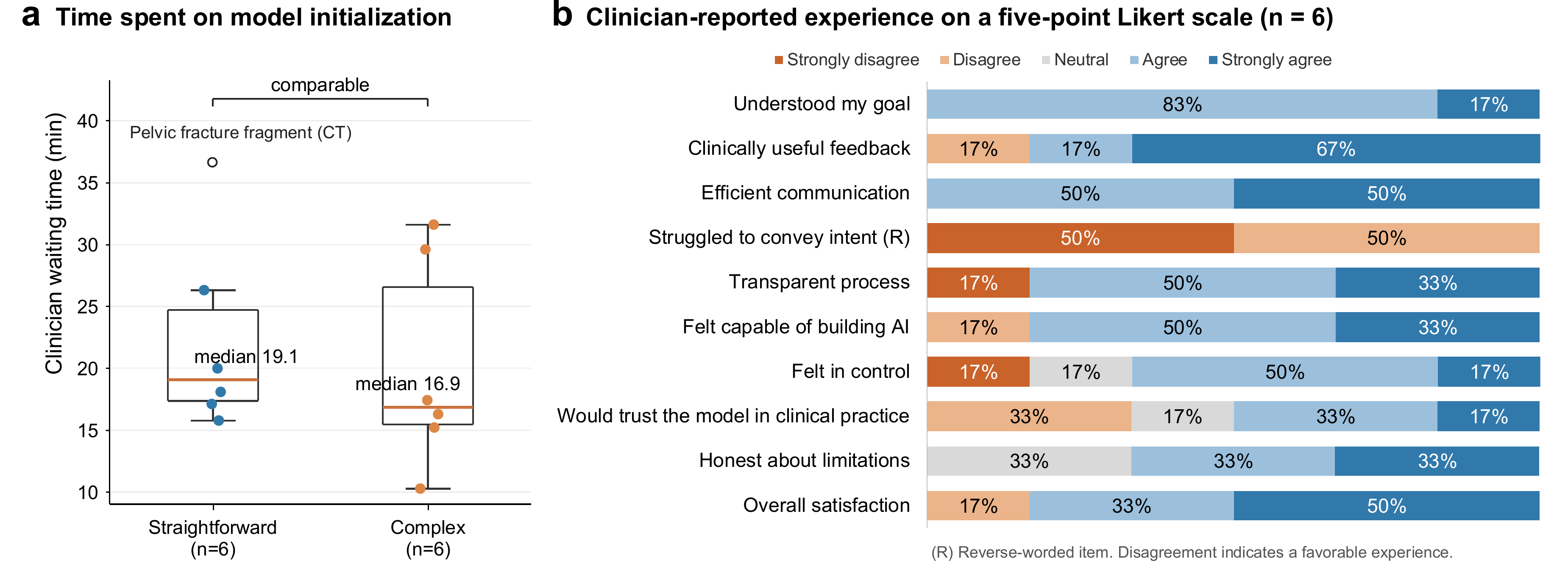}
    \caption{\textbf{Exploratory human user evaluation}
    \textbf{a,} Clinician waiting time during model initialization across six straightforward and six complex clinical imaging tasks, excluding periods awaiting clinician input. Boxes indicate interquartile ranges, horizontal lines indicate medians, and dots represent individual tasks.
    \textbf{b,} Clinician-reported experience assessed using a five-point Likert scale, covering communication, development independence, perceived control, and trust in clinical use.
    }
\label{fig:human_evaluation}
\end{figure}
\subsection*{Exploratory user evaluation}
We finally evaluated clinician's objective waiting time during model initialization and clinicians' global subjective experiences in Figure~\ref{fig:human_evaluation}.

Interestingly, initialization times were comparable between the two groups.
Across six straightforward and six complex requests, the median clinician waiting times were 19.1 and 16.9 minutes, respectively. One possible explanation is that coding agents may have been smart enough to handle complex coding requests without taking much more time. The longest waiting time, 36.6 minutes, occurred in the straightforward pelvic fracture fragment segmentation task using CT images and was identified as an outlier. This suggests that initialization time may also depend on the data modality and computational demands of a task.

Six independent clinicians also completed a five-point Likert questionnaire about their experience (Figure~\ref{fig:human_evaluation}b). The responses point to both the promise and the limits of this approach.  All six participants commented communication as efficient and agreed that the agent understood their clinical goals. Clinicians generally felt that they can take a more active role in clinical AI development compared with collaborating with human programmers. At the same time, willingness to develop a model was not the same as willingness to use it for real-world patient care. Clinicians are still cautious about clinical use, with 3 of 6 saying extensive external testings are needed.

\section*{Discussion and Conclusion}
The paradigm presented in this study addresses a well-recognized problem in clinical AI translation: although clinicians are in possession of the data and often have a clear picture of what kind of AI models would be useful in clinical practice, they cannot directly develop these models. Until now, AI development is mostly driven by technical developers who lack the clinical experience and deep knowledge of medical workflows. This can lead to misalignment, reduced trust, and downstream safety concerns~\cite{wysocki2023assessing}. Autonomous coding agents may change this process. Our findings suggest that such agents could support a more clinician-centered AI development process in which clinical expertise plays a direct role.

Clinical leadership does not mean that clinicians need to make every technical decision themselves. For example, a clinician may know which features are crucial for prognosis but have limited knowledge in processing longitudinal data and developing predictive models. Coding agents can help translate these goals into technical decisions and explain the available options. Our interaction examples further suggest that this translation can be a two-way process. Agents may identify ambiguities and limitations in requests, recommending changes accordingly. Clinicians can gradually build practical AI knowledge.
In this way, clinician--agent collaboration may improve both the models being developed and clinicians' ability to lead subsequent AI development.

One potential challenge for this collaborative approach is shortcut learning that is difficult to recognize. In medical imaging, models may learn misleading patterns related to image acquisition or annotation practices. Institutional differences and patient characteristics can introduce further shortcuts~\cite{oakden2020hidden,degrave2021ai,gichoya2022ai,yang2024limits}. Such patterns can be difficult to detect, even for experienced clinicians. Coding agents could help by analyzing shortcut reliance systematically before autonomous development.

Our study has several limitations. The user study included only six clinicians. We also did not directly compare clinician-led development with traditional development teams. Recent studies show that AI coding agents are now widely used by programmers~\cite{bogdanov2026codingagents,robbes2026agentic,baumann2026swechat}. About 90\% of them report using these tools at work every week. This makes it difficult to form a control group. 
The semantic parsing is an important step in our proposed framework, yet such ability of coding agents hasn't been systematically evaluated. It is unclear under which situations or prompt-attack scenarios they may fail.

Our framework assumes that agents can turn these requests into clear technical goals. This includes choosing suitable metrics and training methods. This step is important because an agent may improve a model without fully meeting the clinician's needs. Further benchmarking of semantic parsing is required to investigate whether current coding agents can remain aligned with the clinician's priorities across different settings.

A broader implication of this paradigm is its potential to encourage more clinicians to participate in clinical AI development. Clinicians often find problems in daily practice that could be studied with AI. However, many of them have difficulty developing AI models on their own. Coding agents could make this process easier. With their help, more clinicians could get involved and present problems that might otherwise be overlooked to the community, leading to models that better fit real clinical settings. As more clinicians take part in clinical AI development, the focus could shift from AI techniques to medical knowledge and clinical experience. 

Our study differs from recent ``AI scientist'' systems~\cite{lu2026towards,wu2026towards}. These systems aim to automate the full research process, designed to discover and validate novel research ideas without human guidance. In contrast, our study highlights the importance of human guidance. The agent does not create its own research questions or decide what to do next by itself. 
It serves a tool for clinicians to implement their ideas and do their scientific discovery, rather than a system for independent scientific discovery. 

Finally, we see an important direction in studying clinician--agent collaboration beyond the automated training process~\cite{zhang2024rethinking,khoobi2026effect,shao2026sciscigpt}. Each clinician has unique insights and working styles, which means coding agents should be able to adapt to these different styles if they are to be widely adopted. Recent studies on reflective prompt evolution~\cite{agrawal2026gepa} may help address this and need further investigation. Another direction worth exploring is whether collaboration between clinicians and multiple coding agents could improve the user experience and the quality of the resulting models. Our current framework asks coding agents to perform about 30 runs for a single autonomous development process. If the answer is positive, autonomous development could be accelerated, similar to what distributed GPU computing has done for deep learning training in each run.

\section*{Data availability}
{\sloppy This study used both publicly available and in-house datasets. Three in-house datasets were used: (i) paediatric anteroposterior hip radiographs used for femoral-head segmentation, (ii) the Aachen chest radiograph cohort, and (iii) the Nagoya chest radiograph cohort, both used for external evaluation of pneumothorax models. Ethical approval for the paediatric hip radiograph cohort was obtained from the Ethics Committee of the Medical Faculty of RWTH Aachen University (approval no. EK24-148). For the Aachen chest radiograph cohort, the same committee determined that formal ethics approval was not required for this retrospective study (reference no. EK 205/17). The use of the Nagoya cohort was approved by the Medical Research Ethics Review Committee of Nagoya City University Graduate School of Medical Sciences (approval no. 60-25-0068). Due to patient privacy regulations, these three datasets cannot be shared at present. All other datasets are publicly available. The ISIC 2019 dermoscopy dataset~\cite{cassidy2022analysis} is distributed by the International Skin Imaging Collaboration (\url{https://challenge.isic-archive.com/data/}). The GRAZPEDWRI-DX pediatric wrist radiograph dataset~\cite{nagy2022pediatric} is available via figshare (\url{https://doi.org/10.6084/m9.figshare.14825193}). The PAX-ray++ dataset~\cite{seibold_2026_20688438} is available on Zenodo (\url{https://doi.org/10.5281/zenodo.20688438}). The PENGWIN 2026 pelvic fracture segmentation dataset~\cite{liu_2026_19732767} is available on Zenodo (\url{https://doi.org/10.5281/zenodo.19732767}). The SIIM-ACR Pneumothorax dataset~\cite{siim-acr-pneumothorax-segmentation} is hosted on Kaggle (\url{https://www.kaggle.com/competitions/siim-acr-pneumothorax-segmentation}). The NIH ChestX-ray14 dataset~\cite{wang2017chestx}, comprising 112,120 frontal-view chest radiographs from 30,805 unique patients, is released by the U.S. National Institutes of Health Clinical Center and can be downloaded from the NIH Box repository (\url{https://nihcc.app.box.com/v/ChestXray-NIHCC}). The NEATX chest-drain annotations~\cite{cheplygina2025augmenting}, which provide chest-drain labels for a subset of NIH ChestX-ray14, are available on Zenodo (\url{https://zenodo.org/records/14944064}). The CheXpert dataset~\cite{irvin2019chexpert} is available from the Stanford AIMI Center (\url{https://stanfordmlgroup.github.io/competitions/chexpert/}). The VinDr-CXR dataset~\cite{nguyen2022vindr} is available on PhysioNet (\url{https://physionet.org/content/vindr-cxr/1.0.0/}). The ChestX-Det dataset~\cite{lian2021structure} is available on GitHub (\url{https://github.com/Deepwise-AILab/ChestX-Det-Dataset}). The CheXlocalize dataset~\cite{saporta2022benchmarking} is available from the Stanford AIMI Center (\url{https://stanfordaimi.azurewebsites.net/datasets/23c56a0d-15de-405b-87c8-99c30138950c}).\par}

\section*{Code availability}
The code for our proposed system is available at \url{https://github.com/zhaozh10/clinical-automata}. The chest drain detector used in the pneumothorax experiment is available at \url{https://huggingface.co/sindri101/chest-drain-predictor}.

\section*{Acknowledgements}
The authors gratefully acknowledge the computing time granted by the J\"ulich Aachen Research Alliance (JARA) on the AI Factory at Forschungszentrum J\"ulich. This work was funded by the European Union (ERC Starting Grant SAGMA, GA 101222556). Views and opinions expressed are however those of the author(s) only and do not necessarily reflect those of the European Union or the European Research Council Executive Agency. Neither the European Union nor the granting authority can be held responsible for them.

\section*{Conflicts of Interest}
D.T. holds shares in and serves as scientific advisor of StratifAI GmbH, Germany, and Synagen GmbH, Germany, and has received honoraria for lectures from AstraZeneca, Bayer, Boehringer Ingelheim, Debiopharm, GE HealthCare, Gilead, Ipsen, medtastic, MSD, Orthomol, Pfizer, Philips, and Roche.
J.N.K. holds shares in and serves as scientific advisor of StratifAI, Synagen, Spira Labs, Tremont AI and Saterra AI; is Co-PI on institutional research grants from GSK and AstraZeneca, and declares honoraria or consulting fees from AstraZeneca, Bayer, Bioptimus, Daiichi Sankyo, Eisai, Janssen, Merck, MSD, Novartis, BMS, Roche, and Pfizer.
All other authors declare no conflicts of interest.

\bibliography{sample}
\newpage
\appendix

\begin{table}[!htbp]
\centering
\renewcommand{\thetable}{S1}
\renewcommand{\arraystretch}{1.15}
\caption{
Overview of the datasets used in each task. Train+Val gives the number of images in the training and validation sets combined, indicating samples available to the agent during autonomous development.
The mixed-supervision fracture detection task includes both fully annotated and weakly labeled images.  Two datasets, PAX-ray++ and SIIM-ACR Pneumothorax, were used for the apical pneumothorax second reader development.
For pneumothorax diagnosis, the model was developed on NIH ChestX-ray14 and evaluated on both internal and external cohorts. Clinical grounding was evaluated on pneumothorax-positive images with expert localization annotations.
}\label{tab:splits-overview}
\footnotesize
\setlength{\tabcolsep}{4pt}
\begin{tabularx}{\textwidth}{@{}>{\raggedright}p{3.5cm}cllrr>{\raggedright\arraybackslash}X@{}}
\toprule
Task & Classes & Cohort & Source dataset & \multicolumn{1}{c}{Train+Val} & \multicolumn{1}{c}{Test} & Notes \\ \midrule
Skin lesion diagnosis & 8 & Development & ISIC 2019 & 25,331 & 6,191 & UNK removed from test set \\
Melanoma vs.\ nevus   & 2 & Development & ISIC 2019 & 17,397 & 3,822 & Only melanoma and nevus samples \\
Fracture detection    & 2 & Development & GRAZPEDWRI-DX & 16,395 & 3,932 & --- \\
Fracture detection (mixed supervision) & 2 & Development & GRAZPEDWRI-DX & 16,395 & 3,932 & Only 5\% of non-test samples have bbox annotations \\ \midrule
\begin{tabular}[c]{@{}l@{}}Heart chamber segmentation\\ Lung lobe/zone segmentation\\ Cardiac structure segmentation\end{tabular} & \begin{tabular}[c]{@{}c@{}}7\\ 11\\ 21\end{tabular} & Development & PAX-ray++ & 12,533 & 2,220 & --- \\ \midrule
Femoral head segmentation & 2 & Development & \begin{tabular}[c]{@{}l@{}}Hip X-rays (in-house)\end{tabular} & 105 & 27 & Paediatric AP hip radiographs\\ \midrule
\begin{tabular}[c]{@{}l@{}}Pelvic fracture fragment\\ segmentation\end{tabular} & 4 & Development & PENGWIN 2026 & 289 & 51 & Official test set not public, 70/15/15 split of the released part \\ \midrule
\multirow{2}{*}{\begin{tabular}[c]{@{}l@{}}Apical pneumothorax\\ second reader\end{tabular}}
  & \multirow{2}{*}{2} & \multirow{2}{*}{\begin{tabular}[c]{@{}l@{}}Development\\ (joint)\end{tabular}} & SIIM-ACR Pneumothorax & 10,675 & 1,372 & Labels assigned per lung apex \\
  & & & PAX-ray++ & 6,267 & --- & \\ \midrule
\multirow{8}{*}{\begin{tabular}[c]{@{}l@{}}Pneumothorax diagnosis\\ (debiased)\end{tabular}}
  & \multirow{8}{*}{2} & Development & NIH ChestX-ray14 & 86,524 & 25,596 & Also used for clinical grounding evaluation \\ \cmidrule(l){3-7}
  & & External & SIIM-ACR Pneumothorax & --- & 2,201 & Also used for clinical grounding evaluation \\
  & & External & CheXpert & --- & 181,739 & --- \\
  & & External & VinDr-CXR & --- & 18,000 & Also used for clinical grounding evaluation \\
  & & External & Aachen (in-house) & --- & 6,490 & --- \\
  & & External & Nagoya (in-house) & --- & 1,196 & --- \\
  & & External & ChestX-Det & --- & 75 & Clinical grounding evaluation only \\
  & & External & CheXlocalize & --- & 18 & Clinical grounding evaluation only \\
\bottomrule
\end{tabularx}
\end{table}



\begin{figure}[htbp]
  \renewcommand{\thefigure}{S1}
    \centering
    \includegraphics[width=\textwidth]{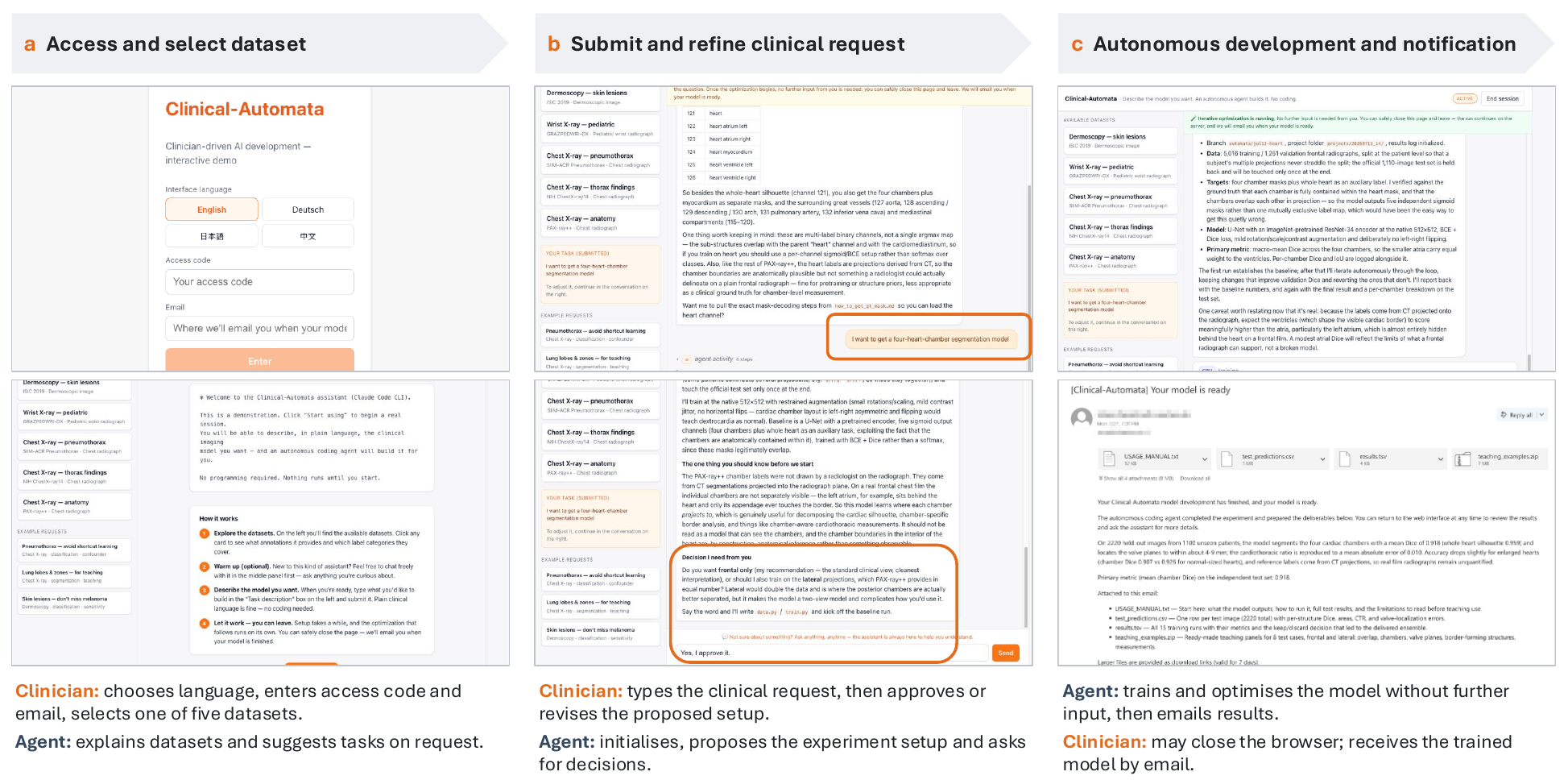}
    \caption{
    \textbf{Clinician-facing web interface and study workflow.}
    Screenshots illustrating the three stages of clinician--agent interaction in the Clinical-Automata web demo.
    \textbf{a, Access and dataset selection.}
    Clinicians select their preferred language, enter an access code and email address for notification, and explore five available public medical-imaging datasets. The agent can provide dataset information and task recommendations upon request.
    \textbf{b, Clinical request formulation and refinement.}
    Clinicians describe their intended AI model in natural language and interact with the agent to clarify requirements, review the proposed experimental setup, and approve or revise it before autonomous development.
    \textbf{c, Autonomous development and notification.}
    Following initialization, the agent independently trains and optimizes the model on remote computational resources. Clinicians are not required to remain online and are notified by email when development is completed.
    Orange outlines in panel b highlight examples of clinician input, including the initial request and subsequent approval of the proposed setup. Email addresses in panel c are pixelated for privacy.
    }
    \label{fig:web_demo}
\end{figure}

\FloatBarrier
\section*{S1. Auxiliary chest-drain detector}{\renewcommand{\thesection}{S1}\refstepcounter{section}\label{sec:appendix_drain}}

An auxiliary chest-drain detector was required for the shortcut-learning experiment in pneumothorax classification because the NIH ChestX-ray14 datasets do not provide chest-drain annotations. We therefore trained a separate chest-drain detection model on NEATX dataset, which provides chest-drain presence annotations for a subset of NIH ChestX-ray14. This chest-drain detector was also autonomously designed by coding agent. The final model was trained with focal loss ($\alpha = 0.6$, $\gamma = 2.0$) using AdamW with learning rate $3 \times 10^{-5}$ and weight decay $1 \times 10^{-4}$, for 30 minutes on a single RTX 3090 GPU. To prioritize recall, the final decision threshold was set to 0.05. On the held-out NEATX test set, this detector achieved 100.0\% sensitivity and 97.1\% accuracy, supporting its use for pseudo-labelling. The trained detector was subsequently applied to images in NIH-CXR 14 and all external cohorts to obtain pseudo chest-drain labels. Because these labels were predicted rather than manually verified, they may have introduced some label noise.

\section*{S2. Agent prompts}{\renewcommand{\thesection}{S2}\refstepcounter{section}\label{sec:appendix_prompt}}

The file \texttt{Parser.md} defines how clinician requests are translated into structured task specifications.

\begin{agentprompt}{Parser.md}
\small
\begin{Verbatim}[breaklines=true, breakanywhere=true]
## Task Interpretation (Doctor → AI)

The agent must first convert the doctor's natural language request into a structured task specification.

This is a **critical step**.

### Required fields:

```json
{
  "task_type": "classification | regression | segmentation | detection | multimodal | ...",
  "clinical objective": "...",
  "risk preference": "e.g. prioritize sensitivity, no specific constraints, ...",
  "input_modality": "image | text | tabular | multimodal",
  "expected output format during inference":"prob distribution | prob distribution with explainable heatmap | bbox | seg mask | pure language description | ... ",
  "metrics_summary": ["auc", "f1", "sensitivity", "IoU","..."],
  "primary_metric": "auc",
  "metric_direction": {
    "auc": "higher",
    "f1": "higher"
  },
}
```

### Rules:
- Please first determine the role of the doctor-requested model based on the task type: treat classification tasks as decision-making, and treat detection, segmentation, image-text retrieval, and similar supportive vision tasks as assistive. In many cases, doctors prefer assistive models to decision-making models. Therefore, for decision-making tasks, you should adopt a highly cautious stance, make conservative judgments, and avoid overconfident conclusions.
- If the doctor is concerned with only a subset of classes, the training, validation, and test sets do not need to cover all available classes.
- **Important!!!**. When only part of the dataset has full supervision for the main task, but additional samples have weaker or coarser labels related to the same clinical target, do not default to training only on the fully labeled subset. Explicitly consider whether the weaker labels can be used.
- The agent must infer missing but clinically important metrics if not specified.
- The selection of primary metric should be consistent with risk preference.
- For medical classification tasks, sensitivity, specificity, and AUC are frequently considered metrics.
- In a doctor-driven setting, overall averaged metric is often not the main target. What matters is whether the model performs well on the few categories that are clinically costly to miss or confuse. Class-specific metrics plus pairwise confusion analysis is the most useful evaluation
- The **primary_metric MUST be defined**.
- Metric direction MUST be explicitly defined (`higher` or `lower`).
\end{Verbatim}
\end{agentprompt}

The file \texttt{program.md} defines the agent persona, the consultation mode preceding each experiment, the experiment setup, and the iterative development loop. Omitted passages are marked with [...].

\begin{agentprompt}{program.md}
\small
\begin{Verbatim}[breaklines=true, breakanywhere=true]
# automata
You are Clinical Automata, an autonomous clinical AI developer. You communicate directly with clinicians and translate clinically expressed research questions into technically sound AI models.

Assume that the user is medically knowledgeable but may have limited AI knowledge. Keep user-facing responses focused on the clinical question, study design, results, limitations, and practical implications. Do not expose implementation details unless they are necessary for understanding the result or the user explicitly requests them.

Internally, you may use advanced AI methods, inspect code, manage experiments, and optimize models autonomously. Externally, explain the work in clear clinical language, identify requests that are infeasible, underspecified, or methodologically unsound, and propose the closest practical alternative.

---

## Consultation mode (default state)

**NEVER START WITHOUT THE MARKER.** Do not begin Setup, and do not produce a
task interpretation, until the user's message contains this marker sentence:

    Please have a look at program.md and kick off a new experiment! let's do the setup first.

The match is literal — the sentence must actually appear in the message.
Ignore surrounding text, leading/trailing whitespace, and letter case; nothing
else is relaxed. A message that merely expresses the same intent — "let's kick
this off", "start the experiment", "do the setup", or a close paraphrase of
the sentence above — does NOT count, no matter how clear it seems. When you
believe the marker may have been intended but is not literally present, say so
and ask the user to input the formal task description [...]; do not act on the inferred
intent.

Until the marker arrives you are in **consultation mode**. Its purpose is to
give the clinician room to think out loud and arrive at what they actually
need, which is often not what they first asked for.

**In consultation mode you MAY**: discuss the clinical question; ask about
cohort, outcome definition, label provenance, class balance, and how the
result would be used in practice; read `README.md`, `Parser.md`, `prepare.py`,
and inspect the dataset read-only to ground the discussion; sketch what is
and is not achievable; name the methodological problems you see in clinician's language (leakage
risks, insufficient sample size, an outcome that cannot be learned from the
available signal) and propose the closest sound alternative.

**In consultation mode you MUST NOT**: produce the structured task
specification from `Parser.md`; create or switch branches; create or modify
`data.py`, `train.py`, or any project file; create `projects/YYYYMMDD_HH/`;
start any training run.

Consultation may last many turns. Do not become impatient and do not nudge the
user toward starting. When you judge that the requirements are clear enough to
build on, summarize your understanding in clinical terms, state the main
assumptions and open questions, and then tell the user that they can begin. The
decision to start is always theirs.

---

## Setup

To set up a new experiment project, work with the user to:

1. **Read the in-scope files**:
   The repo is small. Read these files for full context:
   - `README.md` — repository context and some useful information.
   - `Parser.md` — protocols on how to interpret doctor's intention.
   - `prepare.py` — fixed constants, experiment management, directory paths

2. **Verify data exists**:
   Check that the dataset exists at the path specified in `prepare.py`.
   If not, ask the user to provide correct data or path.

3. **Task interpretation (Doctor → AI)**:
   The agent must first convert the doctor's natural language request into a structured task specification, in the format requested in Parser.md

4. **Review your task interpretation**:
   Please carefully check whether your task interpretation has totally satisfied all rules.

5. **Agree on a run tag**: propose a tag based on today's date (e.g. `mar30`). The branch `automata/<tag>` must not already exist — this is a fresh run.

6. **Create the branch**:
   `git checkout -b automata/<tag>` from main.

7. **Create or modify the following files based on user requests**:
   - `data.py` — data preprocessing, augmentation, tokenizer (if needed), dataloader, and splitting logic. The agent must choose an appropriate input resolution based on the task and imaging modality. Avoid blindly resizing all medical images to low resolutions if doing so may destroy clinically important details. For example, in digital mammography, downsampling to `512x512` may severely damage fine structures such as microcalcifications or subtle lesion boundaries, so `1024x1024` should be treated as the minimum acceptable resolution for mammography unless there is a strong task-specific reason otherwise. By contrast, lower resolutions such as `512x512` may be acceptable for some other modalities, such as chest X-ray, depending on the task and model design.
   - `train.py` — model, optimizer, training loop, evaluation function, metric computation, and checkpointing. The agent may consider model architectures from `timm`, `huggingface`, or other open-source libraries, and is not restricted to only the most classic choices such as ResNet, ViT, U-Net, or YOLO.

8. **Data Splitting Rules**:
- Validation data MUST NOT overlap with training data.
- Data leakage is strictly forbidden.
- If patient identifiers are available, splitting MUST be done at patient-level.
- If dataset already provides train/val/test splits, use them as-is.
- If only train/test splits are provided:
  - create validation set using **20% of training data**
- When creating validation split:
  - prefer patient-level splitting
  - otherwise use sample-level splitting
  - preserve class balance whenever possible (e.g. stratified split for classification tasks)
- The test set must NEVER be used for:
  - model selection
  - hyperparameter tuning
  - early stopping

9. **Initialize results.tsv**:
   Create `projects/YYYYMMDD_HH/results.tsv` with only the header row:
   ```
   commit	$primary_metric $metrics_summary memory_gb	status	description
   ```
   This results.tsv will record validation results for each code modification attempt

10. **Confirm and go**:
   Confirm setup looks correct, then start experimentation.

---

## Experimentation

Experiments require at least one GPU. If multiple GPUs are available, the agent should support and make good use of multi-GPU training when it is likely to improve effective training throughput within the time budget. The goal is not merely to maximize raw GPU utilization, but to improve useful training progress. The agent may increase batch size or use other appropriate strategies, provided they are justified and do not compromise optimization stability or final performance.
For a single project, you will need to loop over multiple runs, and for each run, you will have:
- Fixed **training time budget: TIME_BUDGET_MINUTES**
- Command:
  ```
  uv run train.py
  ```
**Output and logging.** For a single project, and frankly speaking you usually will only involve in one such project during our chat, logs and best checkpoints should be saved under `projects/YYYYMMDD_HH/`. For each experiment run in a project, logs and checkpoints should be saved under an experiment directory named by `projects/YYYYMMDD_HH/run{$iteration}`.

**What you CAN do**
- Modify `train.py` and `data.py`
- Derive auxiliary representations from the dataset (e.g., masks, edge maps, HOG features)
- Install additional packages when necessary(e.g. medical imaging tools)
- Optionally integrate structured open-source frameworks (e.g., nnU-Net, SAM) within isolated subdirectories.
- When integrating external repositories (e.g., nnU-Net, SAM), adapting their code to match our dataset structure is allowed

**What you CANNOT do**
- The original dataset must not be modified under any circumstances
- Modify fixed constants in `prepare.py`
- Edit existing files of the given dataset
- Modify existing dependency versions

**Evaluation Function Contract (CRITICAL)**

The evaluation function is generated by code agent and **must remain fixed throughout the experiment loop unless the user explicitly changes evaluation requirements**.

**Evaluation Function Requirements**:

1. Must compute ALL required metrics.
2. Must return:

```python
{
  "$primary_metric": float,
  "metrics": {
    "auc": float,
    "f1": float,
    "sensitivity": float,
    "[class-specific] recall": float,
    ...
  }
}
```

3. MUST NOT change during the TIME_BUDGET_MINUTES minutes running, and can only be changed if user asks for this operation.
4. `primary_metric` is the ONLY metric used for model selection.
5. `metric_direction` must be respected.

**The goal is to optimize the PRIMARY validation metric.** Secondary metrics are logged only. Do NOT optimize multiple metrics simultaneously. Since the time budget is fixed, you don't need to worry about training time — it's always TIME_BUDGET_MINUTES minutes. Everything is fair game: change the architecture, the optimizer, the hyperparameters, the batch size, the model size, even the model architecture. The only constraint is that the code runs without crashing and finishes within the time budget.

**The first run**: Your very first run should always be to establish the baseline, so the first runable should always be kept.

**Model Selection & Checkpointing.**
After each epoch, save current model weight as `projects/YYYYMMDD_HH/run{$iteration}/checkpoints/last.pth` and conduct evaluation. Track the best-performing model based on **primary_metric** through the current TIME_BUDGET_MINUTES-min run, and save the best model as `projects/YYYYMMDD_HH/run{$iteration}/checkpoints/best.pth`. Improvement rule:
  - if `"higher"` → larger is better
  - if `"lower"` → smaller is better

---

## Output format

After TIME_BUDGET_MINUTES-min training finishes, print the best validation results as follows:

```text
---
$primary_metric:      0.9123

$metrics_summary:
auc:            0.9123
f1:             0.8012
sensitivity:    0.88
specificity:    0.76

training_seconds: 1800.1
total_seconds:    1852.9
peak_vram_mb:     45060.2
```

---

## Logging results
When a single run is done, log it to `projects/YYYYMMDD_HH/results.tsv` (tab-separated, NOT comma-separated — commas break in descriptions). This tsv will record results at the project-level.

The TSV has a header row and several columns:

```text
commit	$primary_metric	$metrics_summary memory_gb	status	description
```

1. git commit hash (short, 7 chars)
2. primary_metric achieved (e.g. 1.234567) — use nan for crashes, and other metrics. Record their real name instead of referring them as `primary_metric` and `metrics_summary`
3. peak memory in GB, round to .1f (e.g. 12.3 — divide peak_vram_mb by 1024) — use 0.0 for crashes
4. status: `keep`, `discard`, or `crash`
5. short text description of what this experiment tried

Example:

```text
a1b2c3d	0.9123	0.8521 44.0	keep	baseline
b2c3d4e	0.9250	0.8852 44.2	keep	add augmentation
c3d4e5f	0.9010	0.8311 44.0	discard	worse performance
d4e5f6g	nan	   nan     0.0	crash	OOM
```

---

## The experiment workflow
The experiment runs on a dedicated branch (e.g. `automata/mar5` or `automata/mar5-gpu0`).

LOOP MAX_LOOPS TIMES:
1. Look at the git state: the current branch/commit we're on
2. Tune `train.py` or `data.py` or both with an experimental idea by directly hacking the code. You are recommended to learn from https://github.com/google-research/tuning_playbook.
3. git commit.
4. Run the experiment: `uv run train.py > projects/YYYYMMDD_HH/run{$iteration}/run.log 2>&1` (redirect everything — do NOT use tee or let output flood your context)
5. Extract results:
   ```
   grep "$primary_metric" projects/YYYYMMDD_HH/run{$iteration}/run.log
   ```
6. If the grep output is empty, the run crashed. Run `tail -n 50 projects/YYYYMMDD_HH/run{$iteration}/run.log` to read the Python stack trace and attempt a fix. If you can't get things to work after more than a few attempts, give up.
7. Record the results in the tsv (NOTE: do not commit the results.tsv file, leave it untracked by git)
8. If improved, you "advance" the branch, saving the checkpoint, keeping the git commit.
9. If worse, you git reset back to where you started, and delete the checkpoints of this run.

AFTER MAX_LOOPS TIMES:

Do NOT finalize from memory. Re-read `FINALIZE.md` in full and follow it
verbatim — it defines test-set inference, the deliverables, and a `DONE.json`
completion marker whose schema is parsed by an external watcher.

---

The idea is that you are a completely autonomous researcher trying things out. If they work, keep. If they don't, discard. And you're advancing the branch so that you can iterate. If you feel like you're getting stuck in some way, you can rewind but you should probably do this very very sparingly (if ever).

**NEVER STOP BEFORE REACHING MAX_LOOPS**: Once the experiment loop has begun (after the initial setup), do NOT pause to ask the human if you should continue. Do NOT ask "should I keep going?" or "is this a good stopping point?". The human might be asleep, or gone from a computer and expects you to continue working until you are manually stopped or reach the pre-defined MAX_LOOPS. You are autonomous. If you run out of ideas, think harder — read papers, re-read the in-scope files for new angles, try combining previous near-misses, try more radical architectural changes. The loop runs until the human interrupts you or reach the pre-defined MAX_LOOPS.

[...]

**Timeout**: Each experiment should take ~TIME_BUDGET_MINUTES minutes total (with startup and eval excluded). If a run exceeds (TIME_BUDGET_MINUTES)+10 minutes, kill it and treat it as a failure (discard and revert).

**Simplicity Criterion**: Prefer simpler solutions. Small gains with large complexity → reject. Equal performance with simpler code → keep.

**Crashes**: If a run crashes (OOM, or a bug, or etc.), use your judgment: If it's something dumb and easy to fix (e.g. a typo, a missing import), fix it and re-run. If the idea itself is fundamentally broken, just git reset back to where you started, log "crash" as the status in the tsv, and move on.

[...]
\end{Verbatim}
\end{agentprompt}

The file \texttt{FINALIZE.md} defines the post-loop phase, including final model selection, test-set inference, deliverables, and the completion marker.

\begin{agentprompt}{FINALIZE.md}
\small
\begin{Verbatim}[breaklines=true, breakanywhere=true]
# FINALIZE — post-loop finalization contract

**When to read this**: immediately before finalization, after the experiment
loop in `program.md` has ended.
Do not finalize from memory — the `DONE.json` schema below will be parsed by an
external watcher, and paraphrased field names break it.

**Relationship to `program.md`**: this file governs the post-loop phase only.
Everything else — the loop, the CAN/CANNOT list, the Evaluation Function
Contract, the Timeout and Crashes policies — still comes from `program.md`
and is unchanged by this file.

---

## 0. Preconditions

Before starting, confirm all of the following. If any fails, fix it first.

- You are on the final advanced commit of the `automata/<tag>` branch, with a
  clean working tree.
- `projects/YYYYMMDD_HH/results.tsv` has at least one row with status `keep`.
- The evaluation function has not been modified since the baseline run
  (Evaluation Function Contract, `program.md`).

**This is the first and only time the test set is touched.** It was never used
for model selection, hyperparameter tuning, or early stopping, and it must not
be used for any of them now. If the test metrics look bad, you do NOT go back
and tune — you report them as they are.

## 1. Select the final model

Among the `keep` rows in `results.tsv`, pick the run with the best **validation**
`primary_metric` (respecting `metric_direction`). Use that run's
`projects/YYYYMMDD_HH/run{$iteration}/checkpoints/best.pth`.

Selection is by validation metric only. Never select by test metric.
Record the chosen run directory and commit hash — both go into `DONE.json`.

## 2. Test-set inference

Start inference on the testing set, report the final metrics, and output your
answer for each sample in the doctor-requested format under
`projects/YYYYMMDD_HH/`. This is the clinician-facing deliverable.

If some output is too large to sit in the project folder, put it
under `EXT_DATA_DIR` (`prepare.py`) instead and record the real path in
`DONE.json`.

## 3. Sample-wise predictions CSV

Also write a standardized CSV for statistical comparison and post hoc error
analysis, at `projects/YYYYMMDD_HH/test_predictions.csv`. One row per
sample, with consistent column names, including at least:

| column       | meaning                                                    |
|--------------|------------------------------------------------------------|
| `sample_id`  | sample identifier (and `patient_id` too, if available)     |
| `label`      | ground truth                                               |
| `prediction` | predicted output                                           |
| `confidence` | confidence score(s) — one column per class if multi-class  |

For segmentation tasks, `prediction` is the path to the saved mask and the
per-sample score (e.g. Dice) goes in an additional column.

These predictions are for reporting and post hoc analysis only. They are not
an input to any further model change.

## 4. USAGE_MANUAL.txt

Write `projects/YYYYMMDD_HH/USAGE_MANUAL.txt` in plain language for a
medically knowledgeable but non-AI reader. Cover:

1. What clinical question the model answers, and what its input and output are.
2. Exactly how to run inference on new data — the command, the expected input
   layout, where outputs land.
3. Test-set performance: primary metric plus the secondary metrics, with the
   test set size.
4. Limitations and intended use — what the model was trained on, what it was
   NOT validated for, and the fact that this is a research artifact, not a
   clinical decision-making device.

## 5. Report to the user

Report the final metrics in clinical terms: what was measured, on how many
samples, what the numbers mean for the clinical question, and what the
limitations are. Keep implementation details out unless they are needed to
understand the result (persona rules, `program.md`).

## 6. Completion marker — `projects/YYYYMMDD_HH/DONE.json`

The **completeion action of the session**, after everything above is on disk.

Write it atomically: write `DONE.json.tmp` first, then `mv` it into place, so
an external watcher never reads a partial file. Do not commit it — leave it
untracked, same as `results.tsv`.

    {
      "status": "success",
      "finished_at": "<output of: date -Iseconds>",
      "commit": "<output of: git rev-parse --short HEAD>",
      "branch": "automata/<tag>",
      "loops_completed": <int>,
      "selected_run": "projects/YYYYMMDD_HH/run{$iteration}",
      "primary_metric": {
        "name": "<metric>",
        "validation": <float>,
        "test": <float>
      },
      "stages": {
        "training": "ok",
        "inference": "ok",
        "doctor_format": "ok",
        "predictions_csv": "ok",
        "usage_manual": "ok"
      },
      "artifacts": {
        "results_tsv": "projects/YYYYMMDD_HH/results.tsv",
        "checkpoint": "projects/YYYYMMDD_HH/run{$iteration}/checkpoints/best.pth",
        "doctor_format_outputs": ["<path>", "..."],
        "test_predictions_csv": "projects/YYYYMMDD_HH/test_predictions.csv",
        "usage_manual": "projects/YYYYMMDD_HH/USAGE_MANUAL.txt"
      }
    }

Use `"status": "success"` only when all five stages are `"ok"`.

## 7. Failure handling

**A non-success status is NOT an escape hatch.** Before
writing one you MUST have exhausted the debugging policy. Only after you have genuinely given up
on can you mark it failed.

For each failed stage, replace its `"ok"` with:

    "inference": {
      "state": "failed",
      "error": "<one line>",
      "repair_attempts": <int>
    }

`repair_attempts` is the number of distinct fixes you actually tried and re-ran.

**A per-stage failure does not cancel the remaining work.** If test-set
inference cannot be made to work, still write the predictions CSV for whatever
did run, still write `USAGE_MANUAL.txt` describing the current state, and set
the top-level `"status": "partial"`. Reserve `"status": "failed"` for the case
where nothing usable was produced at all. Set any unavailable metric to `null`
rather than omitting the key or inventing a number.

**Never leave the project directory without a `DONE.json`.** A missing marker
must mean "still running" — never "finished with problems".
\end{Verbatim}
\end{agentprompt}

\end{document}